\documentclass{llncs}
\usepackage{graphicx}
\usepackage{cite}
\usepackage{url}
\usepackage{amsmath}
\usepackage{algorithm, algorithmic}
\usepackage{listings}
\usepackage[T1]{fontenc}
\usepackage{courier}
\usepackage{verbatim}

\begin{document}

\title{Task Interaction in an HTN Planner}

\author{Il\v{c}e Georgievski \and Alexander Lazovik \and Marco Aiello}
\institute{Distributed Systems Group\\
Johann Bernoulli Institute for Mathematics and Computer Science\\
University of Groningen \\
\email{\{i.georgievski, a.lazovik, m.aiello\}@rug.nl}}

\maketitle

\begin{abstract}
Hierarchical Task Network (HTN) planning uses task decomposition to
plan for an executable sequence of actions as a solution to a
problem. In order to reason effectively, an HTN planner needs
expressive domain knowledge. For instance, a simplified HTN planning
system such as JSHOP2 uses such expressivity and avoids some task interactions due to the increased complexity of the planning process. We address the possibility of simplifying
the domain representation needed for an HTN planner to find good
solutions, especially in real-world domains describing home and
building automation environments. We extend the JSHOP2 planner to
reason about task interaction that happens when task's effects are
already achieved by other tasks. The planner then prunes some of the
redundant searches that can occur due to the planning process's
interleaving nature. We evaluate the original and our improved planner
on two benchmark domains. We show that our planner behaves better by
using simplified domain knowledge and outperforms JSHOP2 in a number
of relevant cases.
\keywords{HTN planning, task interaction, phantomization, domain knowledge}
\end{abstract}

\section{Introduction}
Hierarchical Task Network (HTN) planning is a well-known and widely used
artificial intelligence planning technique. HTN planners are provided with a set
of goal tasks that have to be repeatedly decomposed until primitive tasks are
reached. Such task decomposition is based on the heuristics or subplans
contained in the domain knowledge. Well written heuristics or methods can
significantly reduce search space and help planner to find an efficiently
executable plan. The main advantage of using HTN technique is the way of writing
the heuristics which can be seen as recipes that fit well with human being
thinking. Using heuristics during the planning process means that there should
be different interactions between tasks.

One type of interaction describes how task's effects are already achieved by other task(s) at some place in the task
network. Most of the HTN planners solve this task interaction by using
the notion of a 
{\em phantom task}~\cite{DBLP:conf/ijcai/Tate77}. Among these
planners, a very popular implementation is SHOP2, a sound and complete planner
\cite{DBLP:journals/jair/NauAIKMWY03}. The SHOP2 planner can reason if certain
effect has been achieved only if the domain knowledge contains such a phantom
task (or several phantom tasks). However, this type of reasoning depends on the
domain's writer ability and experiences to identify and encode such task
interaction and not on the planner's reasoning capabilities. Planners as SHOP2
avoid some task interactions because they can increase the complexity of the
planning process.

In general, writing knowledge for controlled/synthetic domains is
easier than for real-world ones. Even more, it is cumbersome to identify interactions between tasks in real-world settings. As an illustration, we introduce GreenerBuildings\footnote{GreenerBuildings is an Information and Communication Technologies project funded under the European Seventh Framwork Programme on Engineering of Networked Monitoring and Control Systems and Wireless Sensor
Networks and Cooperating Objects. More information can be found on
\url{http://greenerbuildings.eu}} project. GreenerBuildings's
objective is to develop generic principles and an energy-aware
framework that utilizes human activity and context recognition
techniques to adapt buildings for energy saving. To date, buildings
involve many manual tasks, as switching lights and appliances to
seasonal changes in the heating systems. Nevertheless, substantial energy savings in buildings can be achieved by globally switching and regulating installations and appliances to actual needs. Hence, the approach introduces concepts for self-powered sensing, processing and actuation in large distributed networks that dynamically minimize energy consumption. Such distributed networks are planned to have approximately 1000 devices and, thus, a very large number of operators (if we assume that each device represents appropriate operator). Encoding such a complex domain is very tedious, especially when optimal solutions are essential as the goal of the GreenerBuildings is to be energy aware. Operators can be associated with costs that reflect the amount of energy their execution requires. Considering this information, the planner has to choose a decomposition that corresponds to the minimum energy consumption. Even more, this decomposition should be optimal i.e. it should not contain redundant operators. Hence, writing phantom tasks is not the best solution. Taking task interactions into account, such a redundancy can be avoided in simple and elegant way.

In this paper, we extend the best-known JSHOP2\footnote{A Java implementation
of SHOP2. JSHOP2 is available as open source on
\url{http://sourceforge.net/projects/shop/files/JSHOP2/}} to reason
explicitly over the above-mentioned task interactions. We  transfer some of the domain expressivity into the planning
process itself with the goal of keeping the domain representation as
simple and compact as
possible. This is especially useful when the domain writer is not familiar with some
specific-purpose representation, e.g., a phantom task. We introduce a task-to-task
matching, which finds a task that is accomplished and reasons that current
task's effects are achieved by already accomplished task and that the current
task can be ignored, if and only if the effects are still holding. By adding the
ability to identify and solve such task interaction, the planner can also
control the search space by avoiding some redundant paths. Moreover, we show that the planner performs very good in cases when JSHOP does not.

The rest of the paper is organized as follows. Section~\ref{bgr} defines the basic HTN terms and the pantomization process. Section~\ref{inter} introduces the algorithm in details and an example of applying it. Next, Section~\ref{impl} shows the implementation and the evaluation on two benchmark domains. Section~\ref{disc} discusses the addressed issue and reviews related work. Finally, we finish with concluding remarks in Section~\ref{conl}.

\section{Background}
\label{bgr}
\subsection{HTN Planning}
Intuitively, an HTN planning technique can be viewed as an extension of the
classical planning approach. The objective of an HTN planner is not to achieve a
set of goals but instead to perform a set of (goal) tasks. An HTN domain
contains, besides the set of operators, a set of high-level descriptions called
methods. Each method can be decomposed into a task network or a set of tasks
with ordering constraints between them. Each task should satisfy certain
conditions. An HTN planning problem includes initial task network that is
decomposed into a sequence of operators i.e. a plan.

\begin{example}
Consider the following example of high-level descriptions. We have two HTN
methods, one for adapting the office for living atmosphere({\em adjust-office})
and another method for adjusting the work desk ({\em adjust-desk}). Task {\em
adjust-office} can be decomposed into task network of three subtasks {\em
set-AC}, {\em turn-on-light}, {\em turn-on-music}. Task {\em adjust-desk}
contains decomposition of two subtasks, namely operators {\em turn-on-light} and
{\em start-computer}.
\end{example}

Next, we provide formalism for an HTN planning that follows the one of Ghallab
{\em et al.}~\cite{DBLP:books/daglib/0014222}.

A {\em task} is an expression formed by a {\em task symbol} and a set of terms
that can be constants or variables. If the task symbol is an operator symbol,
then the task is {\em primitive}; otherwise, the task is {\em nonprimitive}. In
our example, {\em set-AC}, {\em turn-on-light}, {\em turn-on-music}, and {\em
start-computer} are primitive tasks, while {\em adjust-office} and {\em
adjust-desk} are nonprimitive tasks. 

\spnewtheorem{mydef}[theorem]{Definition}{\bfseries}{\normalfont}

\begin{mydef}
A {\em task network} is a pair $w=(U,C)$ where $U$ is a set of task nodes
and $C$ is a set of constraints. Each task node $u \in U$ contains a task $t_u$.
The task network is primitive if all of the tasks are primitive; otherwise, $w$
is nonprimitive.
\end{mydef}

Each constraint in $C$ specifies a condition that must be satisfied by every
plan that is a solution to the planning problem. Examples of constraints are
{\em precedence constraint}, {\em before-constraint}, {\em after-constraint} and
{\em between-constraint}.

Considering the above example, we could have a task network where $U=\{u_1,u_2,u_3\}$, $u_1=$ {\em set-AC}, $u_2 =$ {\em turn-on-light}, and $u_3=$
{\em turn-on-music}, and $C$ contains a precedence constraints, for example,
that $u_1$ must occur before $u_2$ and $u_2$ must occur before $u_3$ and a
before-constraint such that the air conditioning system is serviceable before
$u_1$.

\begin{mydef}
An {\em HTN method m} is a 4-tuple {\em (name(m), task(m), subtasks(m),
constraints(m))} respectively referring to the method's name, a nonprimitive
task, and the method's task network containing subtasks and constraints.
\end{mydef}

The descriptive name for the nonprimitive task {\em adjust-office} could be {\em adjust1} with subtasks and constraints described above.

A method instance $m$ is applicable in a state $s$ if its preconditions are
satisfied in the $s$. A method instance $m$ is relevant to task $t$ if there is
a substitution $\sigma$ such that $\sigma(t)=task(m)$.

\begin{mydef}
An {\em HTN planning domain} is a pair $D=(O,M)$, where $O$ is a set of
operators and $M$ is a set of methods.
\end{mydef}

Taking into account our example, we could define a domain with five operators and two methods.

An operator $o$ is an action described by a 3-tuple {\em (name(o),
preconditions(o), effects(o))} referring to operator's name, preconditions and
effects, respectively. 

\begin{mydef}
An {\em HTN planning problem} is a 4-tuple $P=(s_0,w,O,M)$, where $s_0$
is the initial state, $w$ is the initial task network and pair $(O,M)$ is the
planning domain.
\end{mydef}

Finally, a plan $\pi=\langle$$a_1$,$a_2$,$\dotsc$,$a_n$$\rangle$ is a solution for planning problem $P$ if there is a sequence of task decompositions that can be applied to $w$ to produce a primitive task network $w'$ such that $\pi$ is a solution for $w'$ (considering that $w$ is a nonprimitive).

For example, the plan $\pi=\langle${\em set-AC,turn-on-light,start-computer,turn-on-music}$\rangle$ could be a solution for the problem of achieving both tasks, {\em adjust-office} and {\em adjust-desk}.

\subsection{Phantomization}
A good and optimal plan should not contain redundant primitive tasks. There are
different aspects of how can unnecessary plan steps be reduced. One way to
accomplish reducing is during the planning process considering certain domain
descriptions. 

Tate in~\cite{DBLP:conf/ijcai/Tate77} has introduced the term {\em phantom task}
as a way of treating the task as already achieved at some point in the network
by other task(s). The phantom task can be accomplished by {\bfseries doing
nothing}, if this task is placed in the network at a point where its effect is
still holding. This type od task reduction is known as {\em phantomization}.
Young {\em et al.}~\cite{DBLP:conf/aips/YoungPM94} have stated that the idea of
phantomization is key to the appropriate performance of the planners that
perform task decomposition. The advantage of using phantomization is the planner's ability to reason
which tasks are unnecessary, and, therefore, to produce more efficient plans. The weak point is its identification and encoding into the domain representation.

\section{Task Interaction}
\label{inter}
We address the possibility of diminishing the tedious writing of effective
domain knowledge by introducing an enhanced reasoning over one type of task
interaction and demonstrating it by extending currently the most popular simplified HTN planner JSHOP2. The reasoning is
performed by checking whether the current task's effects are already achieved by
other same named task, and are still holding at the current state. In the case
where these effects are still holding, the planner reasons that this task is
redundant, avoids applying it and continues with the planning process. By
enabling this task interaction, the planner also has to control the search space
as this interaction can happen in different levels of task interleaving which
can lead to redundant searches or plans (if such exist). 

In this way, the planner is enhanced with the ability to find a plan even when
the domain writer does not provide highly efficient (alternative) methods. The
remainder of the section describes the planning algorithm, and a simple problem
example.

\subsection{Algorithm}
Our algorithms outlined in Alg.\,\ref{al1} and Alg.\,\ref{al2} are high-level
descriptions incorporated in the existing JSHOP2 planning and interleaving
algorithms. Algorithm \ref{al1} takes as input a planning problem
($s_0$,$w_0$,$D$), as defined in the previous section, and an agenda $A$.
Algorithm \ref{al2} takes as input a task list $w_i$, and an agenda $A$.

\begin{algorithm}
\caption{task interaction}
\label{al1}
\begin{algorithmic}[1]
\STATE Call \textbf{interleave} to choose an appropriate task $t \in w_i$
\label{call}
\IF {{\em t is a primitive task}}
\label{primitve}
\IF {$t$ is applicable in the current state $s_i$}
\STATE add the $t$'s effects to the $A$
\ENDIF
\label{ifprim}
\ELSIF {{\em t is a nonprimitive task}}
\label{nonprimitive}
\IF {same named method has been previously reduced}
\label{newstep}
\STATE get it's subtasks and call this algorithm recursively
\ENDIF
\label{newstep2}
\IF {$t$ is reducible in the current state $s_i$}
\label{reducible} 
\STATE add $t$ to the reduced methods list
\ENDIF
\ENDIF
\end{algorithmic}	
\end{algorithm}

Let us examine the algorithms in detail. Algorithm \ref{al1} starts with the
interleaving step which calls helper algorithm Alg.\,\ref{al2}, as noted in line \ref{call} (Alg.\,\ref{al2}
shows only incorporated steps in the existing interleaving mechanism of JSHOP2).
Algorithm \ref{al2} prunes each primitive task that is already applied (i.e.~a step in
the potential plan) and its effects are elements of $A$, as it is shown in lines \ref{primitive2}-\ref{applied}. Agenda $A$ contains all facts that are holding up to the $i$-th state. When these conditions are
fulfilled we say that task $t$ is matchable. Formal definitions follow.

\spnewtheorem{mydef2}{Definition}{\bfseries}{\normalfont}

\begin{mydef2}
Let $s_i$ be the current state. Agenda $A$ consists of a set of logical atoms
which values are accurate in the $s_i$ state. 
\end{mydef2}

\begin{mydef2}
Let $t$ be the current primitive task, $t_a$ already applied primitive task,
$s_i$ the current state, and $A$ is the agenda. Task $t$ is matchable with $t_a$
if and only if $t$ and $t_a$ denote same operator instance, and $t_a$'s effects
are elements of $A$ in the state $s_i$. 
\end{mydef2}

\begin{algorithm}
\caption{interleave}
\label{al2}
\begin{algorithmic}[1]
\IF {{\em $t \in w_i$ is a primitive task}}
\label{primitive2}
\IF {same named operator is already applied and $t$'s effects are in $A$}
\STATE $t$ is matchable and do not interleave it
\ENDIF
\ENDIF
\label{applied}
\end{algorithmic}	
\end{algorithm}

Planning continues depending on whether the chosen task is primitive or
nonprimitive. If the chosen task $t$ corresponds to a primitive task, which is
applicable in the state $s_i$, its effects are added to the list of logical
atoms $A$, as noted in lines \ref{primitve}-\ref{ifprim}. When the chosen task $t$ is a nonprimitive task (see line \ref{nonprimitive}), JSHOP2 skips the method's branches for which bindings do not
exist. However, when we consider task interaction we should check also some of
these branches, in particular, those ones that have been already successfully
instantiated (i.e. their primitive subtasks are part of the potential plan).
Therefore, in lines \ref{newstep}-\ref{newstep2} we add the logic to the algorithm that reduces this type of
branch and call the algorithm recursively with branch's subtasks as task list
$w_{i+1}$, which matchability can be further checked. Formal definition follows.

\begin{mydef2}
Let $t$ be the current nonprimitive task, $t_j$ is the j-th branch of $t$, and
$s_i$ is the current state. Task $t$ is reducible into branch $t_j$ if and only
if $t_j$ is instantiated in state $s_k$, where $k<i$.
\end{mydef2} 

JSHOP2 is a partial-ordered planner, which produces all possible combinations of
tasks' sequences. Previously identified task interaction enables additional
interleaving steps between tasks. Indeed, when the interaction has already
happened many of the interleaving steps are not necessary as they produce
redundant searching. Hence, we have added a control ability to the algorithm
that prunes these steps. For instance, if the planner finds a plan at some point
after successful task interaction and backtracks to try other combinations,
without controlling the search it will find a number of plans which are equal as
the first found one.

The algorithm continues searching for other possible branches of $t$ for
which exist appropriate binding in the $i$-th state. Thus, when certain method's
branch is reducible in the current state $s_i$, method's branch is added to the
list of reduced methods (see line \ref{reducible}).

Some of the Alg.\,\ref{al1}'s steps are excluded from the sketch. They are the
same as the those in JSHOP2.

\subsection{An Example}
\begin{figure}
\centering
\includegraphics[width=.9\textwidth]{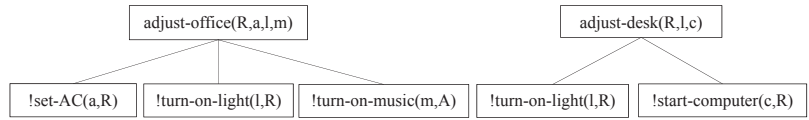}
\caption{Example of two tasks decompositions}
\label{expl}
\end{figure}

Consider the example from Sec.\,2. Say that we want to perform both tasks i.e. two goal tasks: preparing the work desk in office $R$ and adjusting office $R$ after some time being empty. More specifically, the task of adjusting the desk can be decomposed into two subtasks of turning on the light $l$ and starting the computer $c$. Task of adjusting the office can be decomposed into three subtasks of setting the air conditioning system $a$, turning on the light $l$ and switching on the music system $m$ (see Fig.\,\ref{expl}). One of the most effective solutions is when the air conditions are comfortable, the light is turned on, the music is playing and the computer is started and ready for work.

Listing \ref{l1} outlines the above methods described with the JSHOP2 notation.
Methods \texttt{adjust-desk} and \texttt{adjust-office} do not contain any
preconditions to keep representation as simple as possible. Methods'
descriptions are the same as in the graphical design except that they are
generalized for any terms by using variables.

\lstset{basicstyle=\small}

\begin{lstlisting}[frame=single, caption=Simple Methods' Descriptions,
captionpos=b, label=l1]
(:method (adjust-office ?r ?a ?l ?m)
 ()
 ((!set-AC ?a ?r)(!turn-on-light ?l ?r)(!turn-on-music ?m ?r)
 )
)
(:method (adjust-desk ?r ?l ?c)
 ()
 ((!turn-on-light ?l ?r) (!start-computer ?c ?r))
)
\end{lstlisting}

We can now examine the situation when the algorithm is on the right way of
finding a good solution. Let us assume straightforward applying of the first two
operators (\texttt{!set-AC a R}) and (\texttt{!turn-on-light l R}) from method
(\texttt{adjust-office R a l m}). We should note that their effects are added to
the agenda at the applying point. Process continues by interleaving the method
(\texttt{adjust-desk R l c}) and reducing it to its task network. Method's first
subtask is (\texttt{!turn-on-light l R}) which we assumed that is already a part
of the potential plan. Thus, the algorithm reasons that this task is already
achieved and that its effect (the light $l$ is on) is still holding. Therefore,
the algorithm is allowed to prune the task from interleaving and continues by
processing the rest of available tasks. In few steps it finds the correct
sequence of operators, i.e.~the plan.

In contrast to our solution, JSHOP2 will not find a plan
by having in mind the above domain description. In order to be able to find a
solution, we have to improve the domain with more effective descriptions. In
List.\,\ref{l2} we enclose enhanced descriptions. As we can see, we have
included additional method \texttt{light-helper} which has a decomposition
representing a phantomization process of doing nothing when the light is already
turned on. Comparing List.\,\ref{l1} and List.\,\ref{l2} is obvious that we make more simple and compact domain representation.

\begin{lstlisting}[frame=single, caption=Enhanced Methods' Descriptions,
captionpos=b, label=l2]
(:method (adjust-office ?r ?a ?l ?m)
 ()
 ((!set-AC ?a ?r)(light-helper ?l ?r)(!turn-on-music ?m ?r)
 )
)
(:method (adjust-desk ?r ?l ?c)
 ()
 ((light-helper ?l ?r) (!start-computer ?c ?r))
) 
(:method (light-helper ?l ?r)
 (not (on ?l ?r) )
 (!turn-on-light ?l ?r)
 
 (on ?l ?r)
 ()
) 
)
\end{lstlisting}

\section{Implementation and Evaluation}
\label{impl}
As mentioned above, we used SHOP2 Java implementation. We have evaluated our
implementation on two benchmark planning domains, namely the Logistic domain
which is available with the JSHOP2 source code and Dock-Worker
Robot~\cite{DBLP:books/daglib/0014222} which we adapted to fit with the JSHOP2
notation.

We have three objectives by performing evaluation: to show that our enhanced
eJSHOP2 planner is able to find solutions with simplified domain description, to
show that its performance is nevertheless reasonable compared to JSHOP2 planner despite added complexity to the planning process, and to present that our planner can perform better in
some cases.

\begin{figure}
\centering
\includegraphics[width=.9\textwidth]{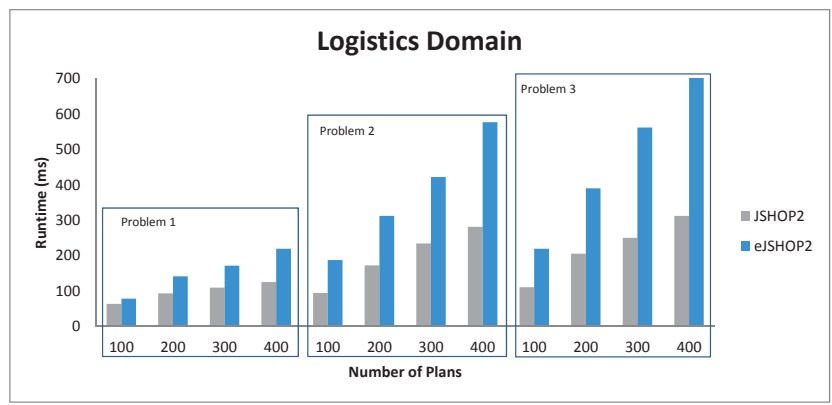}
\caption{Runtime Measurement on the Logistics Domain}
\label{log}
\end{figure}

Figure \ref{log} shows the results by testing the planners in the Logistics
domain. We have created three problems with gradual complexity. Problem 1
includes two locations and two packages that have to be transfered from one
location to the other. Problem 2 examines the situation with three locations and
four packages, and Problem 3 tests 4 locations with six packages to transfer.
For each problem the planners have tried to find 100, 200, 300 and 400 plans. As
we can see, in every case JSHOP2 has better
performance. The reason for such a behavior is due to the reasoning in the
preconditions and not having additional search spaces to look for a plan, while
in our implementation planning includes more chances for additional search space
and backtracking along with it.

\begin{figure}
\centering
\includegraphics[width=.9\textwidth]{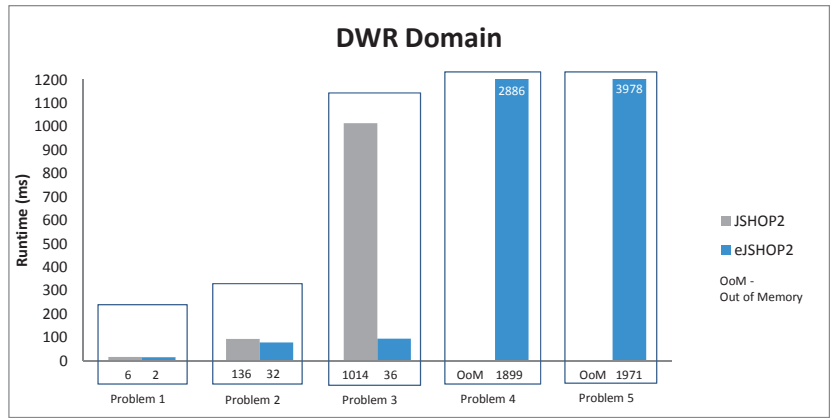}
\caption{Runtime Measurement on the Dock-Worker Robots Domain}
\label{dwr}
\end{figure}

Figure \ref{dwr} shows a little different results. Planners have been tested on
the Dock-Worker Robots domain. There are five different problems on which
planners have tried to find all possible solutions. JSHOP2 planner has found more plans due to its inability to prune the search
space that promises a number of redundant plans as a result of the combinatorial
nature of the interleaving process. In a slightly more complex problems it
easily runs out of memory. On the contrary, our eJSHOP2 planner performs faster
as a result of the possibility to prune some of the ways that can lead to
redundant searches.

All experiments were performed on a Windows-based machine with 3\,GB of memory
and a 2.00\,GHz Intel Core 2 Duo processor.

\section{Discussion and Related Work}
\label{disc}
The main advantage of HTN planning is its ability to deal with very large and
complex problem domains. However, HTN requires experienced domain writers to
provide the planner with the descriptions that it needs to plan. Especially,
this is a case with HTN planners that know everything about the state of the
world at each planning step. JSHOP2 produces a plan based on such a principle
and executes the plan latter. Worthily mentioning is that this kind of planners
guarantee that a plan will be found if all the necessary requirements are
provided. Indeed, we have to write powerful methods and to describe all possible
situations that might happen in the specific domain. Therefore, we have to
assume that we exactly know how these descriptions will affect on the planning
process and the world state. However, we question if there is a possibility to transfer some of the expressivity of the domain representation to the planning process itself.

We have shown that in certain cases the domain representation can be simplified.
Although this contribution has slowed down the performance of the planner, there
are cases when this implementation can be useful. Planners as JSHOP2 avoid task
interactions since they can expand the search space and increase the complexity
of the planning process.

SHOP2 is sound and complete planner~\cite{DBLP:conf/ijcai/NauMCLM01} or it can produce a correct plan if all necessary requirements are fulfilled i.e. a set of methods and operators is able to generate a solution for a problem. Understanding the completeness and soundness of our extension is straightforward as by simplifying the domain representation we enhanced the algorithm to be able to deal with such simplification.

Usually planners' performances are evaluated on standard benchmark domains. However, there is no simple and absolute way to judge the efficiency of a planning system. One should understand the computational complexity of planning in a particular domain in order to be able to asses the efficiency of the planner. If no planning system performs well in a given domain, does it mean that all planners are bad, or is it domain naturally hard? Such questions and answers are addressed in~\cite{DBLP:books/sp/Helmert2008}. 
 
The idea of reducing redundant tasks is not new. Past 30 years a lot of research
has been done in finding a way of reducing unnecessary plan steps. One way to
accomplish this idea is during the planning process. Several task decomposition
planners support this way of
reduction~\cite{Sacerdoti}~\cite{DBLP:conf/ijcai/Tate77}~\cite{Wilkins}. All planners use the domain expressivity to reason about the redundancy and, to the best of our knowledge, no one has incorporated that ability to the planner itself.
 
As mentioned, Tate~\cite{DBLP:conf/ijcai/Tate77} has introduced the
phantom task that can be accomplished by doing nothing, if this task is placed
in the network at a point where its effect is still holding. The phantomization
process is used in the framework for plan modification and
reuse~\cite{DBLP:journals/ai/KambhampatiH92}. It is stated that when the task
$t$ is of the form {\em achieve(C)}, and $C$ can be achieved directly by using
the effects of some other task $t_c \in T$, where T is a set of tasks, then $t$
becomes a phantom task and its reduction becomes $\langle$\{{\em
phantom(C)}\},{\O},{\O}$\rangle$. All these task reduction schemas are given to
the planner {\em a priori} as part of the domain specification.

In addition,~\cite{DBLP:books/daglib/0068932} has proposed an idea to merge
tasks with similar actions. To merge two tasks means that their actions must
``match'' in the sense that they differ only in the slots where one or both have
anonymous constants. This idea has no formalism nor practical implementation, thus, it is impossible to compare.

Foulser {\em et al.}~\cite{DBLP:journals/ai/FoulserLY92} have proposed a
formalism and both optimal and efficient heuristics algorithms for finding
minimum-cost merged plans. In their formal theory a set $\Sigma$ of operators is
mergeable with a (merged) operator $\mu$ if and only if $\mu$ can achieve all the
useful effects of the operators in $\Sigma$, and $\mu$'s preconditions are subsumed
in the preconditions of $\Sigma$, and the cost of $\mu$ is less than the cost of
$\Sigma$. Although our idea is similar to theirs, we use less constrained task to task interaction in an HTN planning. Hence, we consider two types of tasks, methods and operators. 

In the multiagent environment, Cox and Durfee~\cite{DBLP:conf/atal/CoxD03} have described an algorithm that uses a merging approach to help agents remove redundant plan steps while at the same time can preserve their autonomy. Basically, their approach remove redundant steps from plans instead of not adding them at all. 

\section{Concluding Remarks}
\label{conl}

We presented an algorithm with the ability to reason about one type of task
interaction. Most of existing systems use domain's methods to deal with this
kind of interaction. We included the phantomization process as part of the
planning process itself.

We have extended JSHOP2 planner to be able to identify and process
already existing effects. Thus, the planner is capable of pruning some redundant
searches. This was not case with previous systems as this cannot be included as
ability into the domain heuristics.

In future work, there is possibility to extend this task-to-task matching into a
task-to-tasks matching. In this case, it is possible to optimize some plans
even more by reducing the number of steps necessary to have a solution to the
planning problem. We believe that there is an opportunity to include more task
interactions into the planning process to the planners as JSHOP2.

\subsubsection*{Acknowledgment.}
We thank our colleague Eirini Kaldeli for the helpful discussions. This work is supported by the GreenerBuildings Project funded under the European Seventh Framework Programme (FP7) with EC contract number INFSO-ICT-258888.

\bibliographystyle{splncs03}
\bibliography{bib}

\begin{thebibliography}{10}
\providecommand{\url}[1]{\texttt{#1}}
\providecommand{\urlprefix}{URL }

\bibitem{DBLP:books/daglib/0068932}
Charniak, E., McDermott, D.: Introduction to artificial intelligence.
  Addison-Wesley series in computer science, Addison-Wesley (1986)

\bibitem{DBLP:conf/atal/CoxD03}
Cox, J.S., Durfee, E.H.: Discovering and exploiting synergy between
  hierarchical planning agents. In: AAMAS. pp. 281--288 (2003)

\bibitem{DBLP:journals/ai/FoulserLY92}
Foulser, D.E., Li, M., Yang, Q.: Theory and algorithms for plan merging. Artif.
  Intell.  57(2-3),  143--181 (1992)

\bibitem{DBLP:books/daglib/0014222}
Ghallab, M., Nau, D.S., Traverso, P.: Automated planning - theory and practice.
  Elsevier (2004)

\bibitem{DBLP:books/sp/Helmert2008}
Helmert, M.: Understanding Planning Tasks: Domain Complexity and Heuristic
  Decomposition, Lecture Notes in Computer Science, vol. 4929. Springer (2008)

\bibitem{DBLP:journals/ai/KambhampatiH92}
Kambhampati, S., Hendler, J.A.: A validation-structure-based theory of plan
  modification and reuse. Artif. Intell.  55(2),  193--258 (1992)

\bibitem{DBLP:journals/jair/NauAIKMWY03}
Nau, D.S., Au, T.C., Ilghami, O., Kuter, U., Murdock, J.W., Wu, D., Yaman, F.:
  Shop2: An htn planning system. J. Artif. Intell. Res. (JAIR)  20,  379--404
  (2003)

\bibitem{DBLP:conf/ijcai/NauMCLM01}
Nau, D.S., Mu{\~n}oz-Avila, H., Cao, Y., Lotem, A., Mitchell, S.: Total-order
  planning with partially ordered subtasks. In: IJCAI. pp. 425--430 (2001)

\bibitem{Sacerdoti}
Sacerdoti, E.D.: A structure for plans and behavior. Ph.D. thesis, Stanford,
  CA, USA (1975), aAI7605794

\bibitem{DBLP:conf/ijcai/Tate77}
Tate, A.: Generating project networks. In: IJCAI. pp. 888--893 (1977)

\bibitem{Wilkins}
Wilkins, D.E.: Practical planning: extending the classical AI planning
  paradigm. Morgan Kaufmann Publishers Inc., San Francisco, CA, USA (1988)

\bibitem{DBLP:conf/aips/YoungPM94}
Young, R.M., Pollack, M.E., Moore, J.D.: Decomposition and causality in
  partial-order planning. In: AIPS. pp. 188--194 (1994)

\end{thebibliography}

\end{document}